\documentclass{CUP-JNL-DCE}

\usepackage{latexsym}
\usepackage{graphicx}
\usepackage{multicol,multirow}
\usepackage{makecell}
\usepackage{amsmath,amssymb,amsfonts}
\usepackage{mathrsfs}
\usepackage{amsthm}
\usepackage{rotating}
\usepackage{appendix}
\usepackage[authoryear]{natbib}
\usepackage{ifpdf}
\usepackage[T1]{fontenc}
\usepackage{times}
\usepackage{sourcesanspro}
\usepackage{newtxmath}
\usepackage{textcomp}
\usepackage{xcolor}
\usepackage{placeins}
\usepackage{float}
\usepackage{hyperref}

\newcounter{algorithm}
\newcommand{\algcaption}[2]{\par\vspace{0.8\baselineskip}\refstepcounter{algorithm}\noindent\textbf{Algorithm~\thealgorithm. #1}\label{#2}\par\vspace{0.25\baselineskip}}

\articletype{RESEARCH ARTICLE}
\jname{Data-Centric Engineering}
\jyear{2026}
\jdoi{10.1017/dce.2026.xxx}
\copytext{\textcopyright\ The Author(s), 2026. This is an Open Access article, distributed under the terms of the Creative Commons Attribution licence (\url{http://creativecommons.org/licenses/by/4.0/}), which permits unrestricted re-use, distribution, and reproduction in any medium, provided the original work is properly cited.}

\begin{document}

\begin{Frontmatter}

\title[Dynamics-aware equation identification]
{Dynamics-aware identification of governing equations from sparse and noisy data}

\author*[1]{Pongpisit Thanasutives}\email{pongpisit.thanasutives@riken.jp}
\author[1,2]{Yoshinobu Kawahara}
\authormark{Thanasutives}

\address*[1]{\orgdiv{Center for Advanced Intelligence Project (AIP)}, \orgname{RIKEN}, \orgaddress{\city{Osaka}, \country{Japan}}}
\address[2]{\orgdiv{Graduate School of Information Science and Technology}, \orgname{The University of Osaka}, \orgaddress{\city{Osaka}, \country{Japan}}}

\received{DD Month 2026}
\revised{DD Month 2026}
\accepted{DD Month 2026}


\abstract{Sparse identification of nonlinear dynamics (SINDy) and PDE functional identification (PDE-FIND) recover parsimonious ordinary and partial differential equations (ODEs and PDEs) from data. However, sparse and noisy temporal measurements can make derivative estimates unreliable. To address this problem, we evaluate Koopman-based upsampling techniques implemented with dynamic mode decomposition (DMD), extended DMD (EDMD), and optimized DMD. These methods learn finite-dimensional approximations of Koopman evolution on selected observables and are used to interpolate and denoise snapshots inside the observed time window before derivative estimation and sparse regression. The empirical benchmark comprises two ODE systems, Lorenz--63 and Van der Pol, and three periodic PDE systems, Burgers, Fisher--Kolmogorov--Petrovskii--Piskunov (Fisher--KPP), and linear advection--diffusion, over sparse and noisy sampling regimes. Polynomial EDMD gives the strongest ODE results, especially in coefficient accuracy. The PDE results are system-dependent: low-rank DMD-assisted reconstructions improve Burgers and advection--diffusion discovery, while the raw baseline (without upsampling) remains competitive for the Fisher--KPP data. A comparison against linear and smoothing-spline interpolation techniques shows that the selected Koopman-based preprocessors provide overall performance gains over these non-dynamical alternatives. We also demonstrate that DMD-assisted upsampling can stabilize Pareto-based non-oracle support-size selection. Overall, Koopman-based upsampling is best viewed as a dynamics-aware preprocessing step that can reduce derivative-estimation error when its observable representation and low-rank structure are appropriate for the data.}

\begin{policy}[Impact Statement]
Data-driven discovery of governing equations can convert sparse measurement snapshots into interpretable differential equations for scientific modeling, digital twins, and feedback control. In many applications, a critical limitation lies in derivative estimation from noisy, temporally sparse observations. This study evaluates curated upsampling strategies in which finite-dimensional DMD/EDMD models are fitted to observed snapshots and then evaluated at intermediate times before SINDy or PDE-FIND is applied. The benchmarks isolate regimes in which Koopman-based preprocessing improves equation recovery. Polynomial EDMD provides the clearest gains for the tested polynomial ODEs, and low-rank DMD-assisted upsampling provides system-dependent gains in the PDE tests. A comparative study shows that, overall, DMD-assisted upsampling outperforms classical interpolation/smoothing techniques.
\end{policy}

\end{Frontmatter}

\section{Introduction}

Data-driven equation discovery tries to turn time-series measurements into parsimonious differential equations. For ordinary differential equations (ODEs), this means learning a vector field such as $\dot{\mathbf{x}}=\mathbf{f}(\mathbf{x})$ from a measured trajectory. Sparse identification of nonlinear dynamics (SINDy) does this by regressing estimated time derivatives onto a library of candidate functions of the state \citep{Brunton2016}. For partial differential equations (PDEs), the same idea becomes PDE functional identification (PDE-FIND): one estimates $u_t$ from field data and regresses it against candidate terms such as $u$, $u^2$, $u_x$, $uu_x$, and $u_{xx}$ \citep{Rudy2017,Schaeffer2017}. These methods are attractive because the output is an interpretable symbolic equation, not just a black-box predictor. The difficulty is that the regression is only as reliable as the derivative estimates supplied to it. When snapshots are noisy and sparsely sampled in time, finite-difference derivatives can become the main source of error, even if the correct candidate terms are present in the (overcomplete) library. This has motivated robust sparse regression variants, implicit formulations, ensemble procedures, weak-form approaches, and information-criterion-based PDE selection, most of which are implemented as open-source software for sparse model discovery \citep{deSilva2020,Champion2020,Kaheman2020,Messenger2021,Fasel2022,Kaptanoglu2022,Bertsimas2023,Thanasutives2023nPIML,Thanasutives2024UBIC,Thanasutives2025VBIC,Thanasutives2026KOPDE}.

This paper studies a preprocessing step that makes the derivative-estimation problem more robust to noisy, sparse data before SINDy or PDE-FIND is applied. The theoretical starting point is a state-space view of the sampled dynamics. Given a time increment $\tau$, an ODE trajectory or a spatially discretized PDE field can be written as a discrete-time evolution $\mathbf{x}_{k+1}=\mathbf{F}(\mathbf{x}_k)$, where $\mathbf{F}$ is generally nonlinear. Koopman operator theory models the evolution of observables rather than requiring the state update itself to be linear \citep{Koopman1931,Rowley2009,Mezic2013}. For an observable $g$, the Koopman update is $g(\mathbf{x}_{k+1})=g(\mathbf{F}(\mathbf{x}_k))$. For a vector of observables $\boldsymbol{\psi}(\mathbf{x})$, the finite-dimensional approximation used in practice is
\begin{equation}
    \boldsymbol{\psi}(\mathbf{x}_{k+1}) \approx \mathbf{M}\boldsymbol{\psi}(\mathbf{x}_k),
    \label{eq:finite_koopman_update}
\end{equation}
which conceptually explains Koopman learning: the original state dynamics can remain nonlinear, while the evolution of selected observables is approximated by a linear matrix map. Here $\mathbf{M}$ is the column-action matrix on observable vectors. In this work, that matrix map is used to interpolate/insert plausible intermediate snapshots between sparse measurements, so that derivatives are estimated from a smoother and more densely sampled sequence before the sparse-regression stage.

Dynamic mode decomposition (DMD) and its variants like extended dynamic mode decomposition (EDMD) provide concrete data-driven versions of this idea. DMD uses the measured state itself as the observable vector, so it fits a linear map from one snapshot to the next in the original coordinates \citep{Schmid2010,Tu2014,Kutz2016}. State-space DMD provides a direct linear Koopman approximation for temporal interpolation. The main reported preprocessors use EDMD or low-rank DMD variants that are better matched to the nonlinear ODE and high-dimensional PDE settings. EDMD uses a richer observable vector, for example polynomial or radial-basis-function (RBF) features, so the same linear update can represent a broader class of nonlinear state evolution \citep{Williams2015}. Optimized DMD fits a continuous-time low-rank exponential model; the implementation used in the scripts follows the optimized-DMD formulation and uses the PyDMD implementation when available \citep{Askham2018,Demo2018,Ichinaga2024}. In the text and tables we refer to this preprocessor simply as optimized DMD or optDMD. For PDE fields, the state dimension is the number of spatial grid values, so the benchmark also uses proper orthogonal decomposition (POD) to represent each field snapshot by a few dominant spatial modes before fitting the Koopman-based DMD/EDMD update. The abbreviation POD-EDMD-RBF thus means EDMD with RBF observables applied after POD compression. This Koopman-based DMD/EDMD step has a clear purpose: learn short-window snapshot evolution, reconstruct intermediate data, and then hand the reconstructed trajectory to SINDy or PDE-FIND, which performs the equation discovery.

Our research question is practical and straightforward. \textit{Does this Koopman-based interpolation improve the equation recovered by SINDy or PDE-FIND from sparse noisy data?} The answer is, however, not expected to be uniform, because the benefit depends on the observable dictionary, the low-rank structure of the data, and the sampling regime. The benchmark therefore compares a no-upsampling baseline with DMD/EDMD-assisted preprocessors on two ODE systems, Lorenz--63 and Van der Pol, and three periodic PDE systems, Burgers, Fisher--Kolmogorov--Petrovskii--Piskunov (Fisher--KPP), and linear advection--diffusion. It reports support recovery, coefficient error, a clearly defined practical score used only as a synthetic-benchmark diagnostic, and a separate non-oracle model-selection experiment based on Pareto optimization. Algorithm~\ref{alg:workflow}, given below, summarizes the proposed method. The baseline differentiates the sparse noisy samples directly; the assisted methods change only the preprocessing step before derivative estimation and sparse regression.

\vspace{0.6\baselineskip}
\begin{center}
\begin{minipage}{0.94\textwidth}
\hrule
\vspace{0.05\baselineskip}
\algcaption{DMD-assisted upsampling for sparse regression.}{alg:workflow}\vspace{1mm}
\small
\textbf{Input:} Sparse noisy snapshots $\mathbf{Y}=\{\mathbf{y}(t_j)\}_{j=1}^m$, candidate library $\Theta$, and preprocessing method $\mathcal{M}$ with upsampling factor $q$.\\[1mm]
\textbf{1.} If $\mathcal{M}$ is the baseline, set $\widetilde{\mathbf{Y}}=\mathbf{Y}$ and $\widetilde{\mathbf{T}}=\mathbf{T}$.\\
\textbf{2.} Otherwise fit $\mathcal{M}$ (a finite-dimensional DMD, EDMD, optimized-DMD, or POD-EDMD model) to $\mathbf{Y}$ or to POD coordinates of $\mathbf{Y}$.\\
\textbf{3.} Evaluate the learned model at intermediate times within the observed window to obtain $\widetilde{\mathbf{Y}}$.\\
\textbf{4.} Estimate temporal derivatives from $\widetilde{\mathbf{Y}}$; for PDE data, also estimate spatial derivatives from the reconstructed fields.\\
\textbf{5.} Build the SINDy/PDE-FIND library and solve the sequential thresholded least-squares (STLSQ) sparse-regression path.\\
\textbf{6.} For oracle diagnostics, select the threshold using the known synthetic support and coefficients; for non-oracle diagnostics, select along the Pareto curve (e.g., extended Bayesian information criterion (EBIC) vs. support size).\\[1mm]
\textbf{Output:} Discovered sparse equation and benchmark metrics.
\vspace{0.95\baselineskip}
\hrule
\end{minipage}
\end{center}

\section{Methodology}
\label{sec:methods}

In this paper, we use standard linear-algebra notation: $(\cdot)^T$ denotes transpose, $\mathbf{I}$ denotes an identity matrix of compatible size, $\mathbf{X}^{\dagger}$ denotes the Moore--Penrose pseudoinverse of $\mathbf{X}$, $\|\cdot\|_2$ and $\|\cdot\|_F$ denote Euclidean and Frobenius norms, and $\|\cdot\|_0$ counts nonzero entries. When applied to matrices, $\log$ and $\exp$ denote the principal matrix logarithm and the matrix exponential, respectively.

\subsection{DMD-assisted upsampling for sparse regression}

Let the observed ODE snapshots be rows of $\mathbf{Y}\in\mathbb{R}^{m\times d}$ at times $t_1,\ldots,t_m$. A preprocessing method produces an upsampled sequence $\widetilde{\mathbf{Y}}\in\mathbb{R}^{\tilde{m}\times d}$ at times $\widetilde{t}_1,\ldots,\widetilde{t}_{\tilde{m}}$, where $\tilde{m}>m$. For non-baseline methods, derivatives are estimated after this step. Thus the SINDy regression uses
\begin{equation}
    \dot{\widetilde{\mathbf{Y}}} \approx \mathbf{\Theta}(\widetilde{\mathbf{Y}})\mathbf{\Xi},
    \label{eq:ode_sindy_after_upsampling}
\end{equation}
rather than regressing directly on the sparse noisy observations. For PDE data, the field values $\widetilde{u}(\widetilde{t}_j,x_\ell)$ are flattened over time and space before solving
\begin{equation}
    \widetilde{\mathbf{u}}_t \approx \mathbf{\Theta}(\widetilde{u},\widetilde{u}_x,\widetilde{u}_{xx},\ldots)\boldsymbol{\xi}.
    \label{eq:pde_find_after_upsampling}
\end{equation}
Finite-dimensional Koopman-based DMD/EDMD models are used to regularize temporal sampling before the sparse regression stage, which performs the final equation discovery.

\subsection{Sequential thresholded least-squares regression}

For ODEs, we construct a polynomial library
\begin{equation*}
    \mathbf{\Theta}(\widetilde{\mathbf{Y}}) = [\mathbf{1},\widetilde{\mathbf{y}}_1,\ldots,\widetilde{\mathbf{y}}_d,\widetilde{\mathbf{y}}_1^2,\widetilde{\mathbf{y}}_1\widetilde{\mathbf{y}}_2,\ldots],
\end{equation*}
where $\widetilde{\mathbf{y}}_i$ denotes the vector of samples of the $i$th state variable and powers/products are evaluated componentwise over rows. For PDEs, the library is
\begin{equation*}
\mathbf{\Theta} = [\mathbf{1},\mathbf{u},\mathbf{u}^2,\mathbf{u}_x,\mathbf{u}\mathbf{u}_x,\mathbf{u}^2\mathbf{u}_x,\mathbf{u}_{xx},\mathbf{u}\mathbf{u}_{xx}],
\end{equation*}
where powers and products of sampled field vectors are interpreted componentwise after the spatiotemporal samples have been flattened. This library contains the true terms for periodic Burgers, Fisher--KPP, and the advection--diffusion system.

Coefficients are estimated by sequential thresholded least squares (STLSQ). For a single target equation, let $\boldsymbol{d}$ denote the derivative target and let $\boldsymbol{\theta}_j$ denote the $j$th library column. In the ODE case, $\boldsymbol{d}$ is one column of $\dot{\widetilde{\mathbf{Y}}}$; in the PDE case, $\boldsymbol{d}$ is the flattened vector $\widetilde{\mathbf{u}}_t$. Thus the left-hand side of the sparse linear system is a temporal derivative, not the state trajectory itself. STLSQ starts with a ridge estimate
\begin{equation}
    \boldsymbol{\xi}^{(0)} = \arg\min_{\boldsymbol{\xi}} \|\mathbf{\Theta}\boldsymbol{\xi}-\boldsymbol{d}\|_2^2 + \alpha \|\boldsymbol{\xi}\|_2^2 .
    \label{eq:stlsq_initial_ridge}
\end{equation}
At iteration $s$, the active/support set is
\begin{equation}
    \mathcal{S}^{(s)}(\lambda)=\{j: |\xi_j^{(s)}|\geq \lambda\},
    \label{eq:stlsq_active_set}
\end{equation}
and the coefficients are refit on the active columns:
\begin{equation}
    \boldsymbol{\xi}^{(s+1)}_{\mathcal{S}^{(s)}} = \arg\min_{\boldsymbol{\eta}} \|\mathbf{\Theta}_{\mathcal{S}^{(s)}}\boldsymbol{\eta}-\boldsymbol{d}\|_2^2 + \alpha\|\boldsymbol{\eta}\|_2^2,
    \qquad
    \xi_j^{(s+1)}=0 \quad \text{for } j\notin\mathcal{S}^{(s)} .
    \label{eq:stlsq_refit}
\end{equation}
For the multi-output ODE regression in Eq.~\eqref{eq:ode_sindy_after_upsampling}, this procedure is applied columnwise to $\mathbf{\Xi}$. In the implementation, the columns of $\mathbf{\Theta}$ are normalized to unit norm only to condition the ridge solves; the threshold $\lambda$ in Eq.~\eqref{eq:stlsq_active_set} acts on the coefficient magnitudes in physical units, so $\xi_j^{(s)}$ there denotes the physical-unit coefficient, and the returned coefficients are in physical units. Unless otherwise stated, STLSQ uses ridge parameter $\alpha=10^{-8}$ and at most 12 threshold/refit iterations. The ODE threshold grid is $\Lambda_{\rm ODE}=\{0,10^{-4},3\!\times\!10^{-4},10^{-3},3\!\times\!10^{-3},10^{-2},3\!\times\!10^{-2},10^{-1},3\!\times\!10^{-1},1\}$, and the PDE threshold grid is $\Lambda_{\rm PDE}=\{0,10^{-5},3\!\times\!10^{-5},10^{-4},3\!\times\!10^{-4},10^{-3},2\!\times\!10^{-3},3\!\times\!10^{-3},5\!\times\!10^{-3},7\!\times\!10^{-3},10^{-2},1.3\!\times\!10^{-2},1.5\!\times\!10^{-2},2\!\times\!10^{-2},3\!\times\!10^{-2},5\!\times\!10^{-2},7\!\times\!10^{-2},10^{-1},2\!\times\!10^{-1},3\!\times\!10^{-1}\}$.

For each candidate threshold, the benchmark selects the value maximizing the practical score
\begin{equation}
    S(\lambda)=\frac{F_1(\lambda)}{1+E_{\rm coeff}(\lambda)},
    \qquad
    E_{\rm coeff}(\lambda)=\frac{\|\widehat{\mathbf{\Xi}}_\lambda-\mathbf{\Xi}_\star\|_F}{\|\mathbf{\Xi}_\star\|_F},
    \label{eq:practical_score_threshold}
\end{equation}
with the analogous vector form for PDEs. This score is a simple synthetic-benchmark diagnostic that is available only when the true active terms and coefficients are known. Its role is to summarize two familiar quantities, support recovery and coefficient accuracy, in one scalar for threshold-grid comparisons. The support F1 score rewards the correct active terms but ignores coefficient magnitudes, whereas coefficient error can look acceptable for a dense model that fits the dominant coefficients while adding small false-positive terms. Conversely, a model with the right support can still be scientifically poor if its coefficients are biased. Dividing $F_1$ by $1+E_{\rm coeff}$ preserves the support-recovery scale, gives $S=1$ only when the support is exact and the coefficient error is zero, and monotonically penalizes coefficient bias without allowing coefficient accuracy to compensate for a completely wrong support. This makes $S$ useful for the narrow question addressed here: whether a preprocessing method gives STLSQ a cleaner regression problem when the synthetic ground truth is available. It should not be confused with information criteria (e.g., AIC, BIC, and EBIC), cross-validation error, or trajectory-forecast validation, which are model-selection or validation tools that do not assume known coefficients. When tables report Score, $S$ is computed for each evaluation record and then averaged; it should not be recomputed from the displayed mean F1 and median coefficient error. Thus the reported F1 score, coefficient error, and practical score are best-case-over-grid quantities evaluated at the $S$-maximizing threshold, not out-of-sample metrics. Ties are broken by larger F1 score and then by smaller coefficient error. This oracle tuning is used only because it is convenient to isolate preprocessing effects when the true support is known. A threshold grid cannot in general guarantee the true support size or the true support itself; if no threshold yields the correct active set, the support-F1 score records the resulting mismatch.

If the true support size $k_\star$ is available, one could instead use oracle best-subset regression,
\begin{equation}
    \min_{\boldsymbol{\xi}} \|\mathbf{\Theta}\boldsymbol{\xi}-\boldsymbol{d}\|_2^2
    \quad \text{subject to} \quad \|\boldsymbol{\xi}\|_0=k_\star .
    \label{eq:oracle_best_subset}
\end{equation}
This is a valid diagnostic for synthetic studies, but it is not available in real applications and is combinatorial for larger libraries. For experimental data, threshold selection should be replaced by validation residuals, information criteria, or stability selection.

Throughout the manuscript, an \emph{oracle} result means that the threshold or model is selected using information that is available only in synthetic benchmarks, namely the true active library terms and, for the practical score, the true coefficients. These oracle results are therefore controlled diagnostics for comparing preprocessing methods, not operational model-selection procedures. A \emph{non-oracle} result means that the selected model is chosen without accessing the true support or true coefficients; ground truth is used only after selection to evaluate the outcome.

\subsection{Finite-dimensional Koopman approximations via DMD models}

The DMD family used here is the computational procedure by which Koopman-based preprocessing is implemented. Standard DMD forms snapshot matrices
\begin{equation}
    \mathbf{X}=[\mathbf{x}_1,\ldots,\mathbf{x}_{m-1}], \qquad
    \mathbf{X}'=[\mathbf{x}_2,\ldots,\mathbf{x}_m],
    \label{eq:dmd_snapshot_matrices}
\end{equation}
and fits the least-squares map
\begin{equation}
    \mathbf{A}_{\rm DMD}=\mathbf{X}'\mathbf{X}^{\dagger} .
    \label{eq:dmd_least_squares_map}
\end{equation}
Here $\mathbf{X}^{\dagger}$ is the Moore--Penrose pseudoinverse, so $\mathbf{A}_{\rm DMD}$ is the minimum-norm least-squares solution to $\min_{\mathbf{A}}\|\mathbf{X}'-\mathbf{A}\mathbf{X}\|_F^2$ when the solution is not unique. A continuous-time interpolation can be obtained from $\mathbf{L}=\log(\mathbf{A}_{\rm DMD})/\Delta T$, where $\Delta T$ is the observed inter-snapshot spacing, and $\mathbf{x}(t)\approx \exp(\mathbf{L}(t-t_j))\mathbf{x}_j$ between observations. In the implementation, the generator and step matrices are projected to their real parts, e.g. $\mathbf{L}=\operatorname{Re}(\log(\mathbf{A}_{\rm DMD}))/\Delta T$ and $\operatorname{Re}(\exp(\mathbf{L}\delta))$ for a substep $\delta$, matching the real-valued data convention used throughout the benchmark. This is computationally attractive because it is linear in the measured state.

EDMD replaces the state with an observable vector $\boldsymbol{\psi}(\mathbf{x})=[\psi_1(\mathbf{x}),\ldots,\psi_p(\mathbf{x})]^T$. In the implementation and notation below, observable snapshots are stacked as rows,
\begin{equation}
    \mathbf{K}=(\mathbf{\Psi}_{\mathbf{X}}^T\mathbf{\Psi}_{\mathbf{X}}+
    \alpha_{\rm K}\mathbf{I})^{-1}\mathbf{\Psi}_{\mathbf{X}}^T\mathbf{\Psi}_{\mathbf{Y}},
    \qquad
    \mathbf{\Psi}_{\mathbf{X}}=
    \begin{bmatrix}
    \boldsymbol{\psi}(\mathbf{x}_1)^T\\ \vdots\\ \boldsymbol{\psi}(\mathbf{x}_{m-1})^T
    \end{bmatrix},
    \quad
    \mathbf{\Psi}_{\mathbf{Y}}=
    \begin{bmatrix}
    \boldsymbol{\psi}(\mathbf{x}_2)^T\\ \vdots\\ \boldsymbol{\psi}(\mathbf{x}_{m})^T
    \end{bmatrix}.
    \label{eq:edmd_row_normal_equations}
\end{equation}
Thus $\boldsymbol{\psi}(\mathbf{x}_j)^T\mathbf{K}\approx \boldsymbol{\psi}(\mathbf{x}_{j+1})^T$ in row form, or equivalently $\mathbf{K}^T\boldsymbol{\psi}(\mathbf{x}_j)\approx \boldsymbol{\psi}(\mathbf{x}_{j+1})$ in column-vector form. In the code, the EDMD and POD-EDMD normal-equation fits use the small ridge value $\alpha_{\rm K}=10^{-8}$ to avoid ill-conditioned solves; state DMD and POD-DMD use the Moore--Penrose pseudoinverse, implemented either directly or through its rank-truncated SVD form. The ODE benchmark uses polynomial and radial-basis-function dictionaries. The RBF dictionaries used in the code are augmented with the constant observable and the linear state or POD-coordinate terms; this augmentation is what makes the linear state-extraction map $\mathbf{C}$ in the reconstruction formulas below well defined. EDMD is expected to be stronger than state DMD when the governing equations are nonlinear and the dictionary contains the relevant observables.

For PDEs, direct DMD on the full spatial grid is high-dimensional. We therefore first compute a POD approximation
\begin{equation}
    \mathbf{u}(t_j) \approx \bar{\mathbf{u}} + \mathbf{V}_r\mathbf{z}_j,
    \label{eq:pod_reconstruction}
\end{equation}
where $\mathbf{V}_r$ contains the first $r$ right singular vectors of the centred snapshot matrix whose rows are $\mathbf{u}(t_j)^T-\bar{\mathbf{u}}^T$. Equivalently, if the centred snapshot matrix is decomposed by the singular value decomposition as $\mathbf{U}_{\rm c}=\mathbf{P}\boldsymbol{\Sigma}\mathbf{V}^T$, then $\mathbf{V}_r$ is formed from the first $r$ columns of $\mathbf{V}$. These modes are the orthonormal spatial directions that capture the largest snapshot variance, and the corresponding latent coordinate is $\mathbf{z}_j=\mathbf{V}_r^T(\mathbf{u}(t_j)-\bar{\mathbf{u}})$. Thus Eq.~\eqref{eq:pod_reconstruction} is the rank-$r$ SVD projection of a field snapshot: it removes components outside the leading POD subspace and lets DMD/EDMD operate on a low-dimensional trajectory. Throughout the manuscript, observed ODE/field snapshot arrays are row-stacked unless a column-snapshot convention is stated explicitly. EDMD is fit in row form and converted to column action through $\mathbf{M}=\mathbf{K}^T$, whereas optDMD below follows the standard column-snapshot convention.

For optimized DMD, the latent snapshots are collected as columns in $\mathbf{Z}=[\mathbf{z}_1,\ldots,\mathbf{z}_m]$. Rather than first fitting a one-step map and then converting it to continuous time, optimized DMD fits the continuous-time exponential representation directly to all observed snapshots:
\begin{equation}
    \mathbf{Z} \approx \mathbf{B}\mathbf{\Phi}(\boldsymbol{\omega},\mathbf{T}),
    \qquad
    \Phi_{k j}=\exp(\omega_k t_j),
    \label{eq:optimized_dmd_model}
\end{equation}
by minimizing
\begin{equation}
    \min_{\mathbf{B},\boldsymbol{\omega}} \|\mathbf{Z}-\mathbf{B}\mathbf{\Phi}(\boldsymbol{\omega},\mathbf{T})\|_F^2 .
    \label{eq:optimized_dmd_objective}
\end{equation}
Here the columns of $\mathbf{B}$ are spatial coefficients in the POD-coordinate space and $\omega_k$ are continuous-time growth or decay rates. The number of exponentials is at most $r$; it is capped by the number of available snapshots. As a nonlinear observable-lifted DMD alternative for PDEs, POD-EDMD-RBF applies EDMD to RBF observables of the latent coordinates $\mathbf{z}_j$. The RBF-augmented observable vector is first mapped back to its linear POD-coordinate block and then reconstructed as $\bar{\mathbf{u}}+\mathbf{V}_r\mathbf{z}$. The PDE preprocessors therefore combine temporal interpolation with rank-$r$ spatial POD denoising; the benchmark should not be interpreted as isolating temporal upsampling from spatial low-rank filtering in the main PDE tables. Appendix~\ref{app:qr_sensitivity} includes a $q=1$ POD-only control to separate POD denoising from temporal upsampling in the representative sensitivity setting.

The main experiments fix the interpolation factor at $q=5$. Each observed interval is divided into five subintervals before derivative estimation. ODE EDMD uses no POD rank. The main PDE experiments use system-specific ranks $r=8$ for Burgers, $r=2$ for Fisher--KPP, and $r=4$ for advection--diffusion; the non-oracle model-selection experiment uses the same $q=5$ and uses $r=8$ for Burgers and $r=2$ for the Fisher--KPP case with a steep periodic front-type initial condition. These values are fixed computational hyperparameters rather than quantities optimized separately for every noise, sparsity, or seed combination. Appendix~\ref{app:qr_sensitivity} reports a representative $q$--$r$ sensitivity analysis, and Appendix~\ref{app:upsampling_strategies} compares interpolation strategies at fixed $q$ and $r$. The sensitivity results show that the chosen ranks and upsampling factor used in the main study are competitive rather than unstable choices, but they are not claimed to be globally optimal.

\subsection{Temporal interpolation from DMD models}

The upsampling step is restricted to interpolation inside the observed time window $[t_1,t_m]$; no results in this paper require forecasting beyond the last observation. Let the sparse observations be available at uniformly spaced times $t_j=t_1+(j-1)\Delta T$, and let the upsampling factor be $q$. The inserted times are
\begin{equation}
    \widetilde{t}_{j,\ell}=t_j+\frac{\ell}{q}\Delta T,
    \qquad \ell=0,1,\ldots,q-1,
    \label{eq:inserted_times}
\end{equation}
with $t_m$ appended at the end. With the row-stacked EDMD convention above, let $\mathbf{M}=\mathbf{K}^T$ denote the corresponding column-action matrix on observable vectors. For POD-coordinate methods whose fitted map is already written in column form, $\mathbf{M}$ denotes that fitted column-action matrix. Fractional evolution is computed by
\begin{equation}
    \mathbf{M}^{\tau}=\exp\{\tau\log(\mathbf{M})\},
    \qquad 0\leq \tau \leq 1 .
    \label{eq:fractional_matrix_power}
\end{equation}
In the implementation, this fractional power is evaluated by eigendecomposition with a pseudoinverse of the eigenvector matrix and eigenvalues clamped away from zero; for diagonalizable $\mathbf{M}$ this coincides with Eq.~\eqref{eq:fractional_matrix_power} under the principal branch. The resulting step matrices and reconstructed trajectories are projected to real-valued form for the real-valued ODE and PDE data considered here. The principal strategy used for EDMD and POD-EDMD-RBF in the benchmark is a \emph{local-reset interpolation}. Let $\mathbf{z}$ denote the coordinate actually advanced by the fitted map: a state vector for state DMD, an observable vector for EDMD, a POD-coordinate vector for POD-DMD or optimized DMD, or the RBF-augmented observable vector of the POD coordinates for POD-EDMD-RBF. Let $\mathbf{c}_0+\mathbf{C}\mathbf{z}$ map this evolved coordinate back to the physical snapshot. Here $\mathbf{c}_0=\mathbf{0}$ and $\mathbf{C}=\mathbf{I}$ for state-coordinate DMD; $\mathbf{c}_0=\mathbf{0}$ and $\mathbf{C}$ extracts the state components from the EDMD observable vector for ODE EDMD; $\mathbf{c}_0=\bar{\mathbf{u}}$ and $\mathbf{C}=\mathbf{V}_r$ for POD-DMD and optimized DMD, whose fitted maps advance POD coordinates directly; and $\mathbf{c}_0=\bar{\mathbf{u}}$ and $\mathbf{C}=\mathbf{V}_r\mathbf{C}_{\rm ext}$ for POD-EDMD-RBF, where $\mathbf{C}_{\rm ext}$ extracts the linear POD-coordinate block from the RBF-augmented observable vector. Then
\begin{equation}
    \widetilde{\mathbf{z}}_{j,\ell}=\mathbf{M}^{\ell/q}\mathbf{z}_j,
    \qquad
    \widetilde{\mathbf{y}}_{j,\ell}=\mathbf{c}_0+\mathbf{C}\widetilde{\mathbf{z}}_{j,\ell}.
    \label{eq:local_reset_interpolation}
\end{equation}
At the next observed time, the interpolation is reset to the measured snapshot $\mathbf{z}_{j+1}$ before the next interval is filled. This reset makes the procedure an interpolation and denoising preprocessor rather than a long-horizon DMD forecast. It also limits the accumulation of model error over many sparse intervals.

For optimized DMD in the PDE experiments, the fitted continuous-time exponential model is instead evaluated directly on the grid of inserted times within $[t_1,t_m]$:
\begin{equation}
    \widetilde{\mathbf{z}}(t)=\sum_{k=1}^{n_{\exp}} \mathbf{b}_k \exp(\omega_k t),
    \qquad t\in\{\widetilde{t}_i\} .
    \label{eq:optimized_dmd_interpolation}
\end{equation}
This is a global continuous-time interpolation in POD coordinates, not an extrapolative forecast beyond the available data, with $n_{\exp}\leq r$.

Several alternative uses of the same fitted map are possible. A \emph{global rollout} uses
\begin{equation}
    \widehat{\mathbf{z}}_{\rm glob}(\widetilde{t})=\mathbf{M}^{(\widetilde{t}-t_1)/\Delta T}\mathbf{z}_1,
    \qquad
    \widehat{\mathbf{y}}_{\rm glob}(\widetilde{t})=\mathbf{c}_0+\mathbf{C}\widehat{\mathbf{z}}_{\rm glob}(\widetilde{t}),
    \label{eq:global_rollout}
\end{equation}
which is simple but can accumulate drift. The same reconstruction map $\mathbf{c}_0+\mathbf{C}\mathbf{z}$ is used here, so $\widehat{\mathbf{y}}_{\rm glob}$ is always expressed in physical state or field coordinates. A \emph{residual-corrected rollout} adds a smooth correction $\mathbf{r}(t)$ satisfying $\mathbf{r}(t_j)=\mathbf{y}_j-\widehat{\mathbf{y}}_{\rm glob}(t_j)$, where $\mathbf{y}_j$ is the observed physical snapshot at the anchor time $t_j$. The corrected reconstruction is $\widehat{\mathbf{y}}_{\rm glob}(t)+\mathbf{r}(t)$. A \emph{two-sided interpolation} blends forward evolution from the left endpoint with backward evolution from the right endpoint,
\begin{equation}
    \widetilde{\mathbf{y}}_{j,\ell}^{\rm 2s}=\mathbf{c}_0+(1-\tau)\mathbf{C}\mathbf{M}^{\tau}\mathbf{z}_j
    +\tau\mathbf{C}\mathbf{M}^{-(1-\tau)}\mathbf{z}_{j+1},
    \qquad \tau=\ell/q .
    \label{eq:two_sided_interpolation}
\end{equation}
These alternatives were tested in an ablation study in Section~\ref{subsec:strategy_ablation}.

\subsection{Pareto-based model selection for dynamical-systems discovery}

The main benchmark uses oracle support information to isolate the effect of preprocessing. The accompanying code also includes a non-oracle, equation-wise model-selection experiment. This experiment follows the SINDy model-selection practice (suitable for real-world use cases) of constructing a Pareto path of sparse-regression solutions and then applying an information criterion to a much smaller set of candidate models \citep{Akaike1974,Schwarz1978,Mangan2017,Thanasutives2024UBIC,Thanasutives2025VBIC}. For each target equation, we form an STLSQ threshold path of candidate coefficient vectors $\widehat{\boldsymbol{\xi}}_\lambda$ and compute the derivative-regression residual sum of squares
\begin{equation}
    {\rm RSS}_\lambda=\|\mathbf{\Theta}\widehat{\boldsymbol{\xi}}_\lambda-\boldsymbol{d}\|_2^2.
    \label{eq:derivative_rss}
\end{equation}
The vector $\boldsymbol{d}$ is the same temporal-derivative target used in STLSQ: $\dot{\widetilde{\mathbf{Y}}}$ for an ODE target equation or the flattened $\widetilde{\mathbf{u}}_t$ vector for a PDE target. Therefore the EBIC calculation measures derivative-regression fit; it does not require simulating an ODE or PDE trajectory. For a candidate with support size $k_\lambda$ selected from a library of $p$ terms, the EBIC is computed as
\begin{equation}
    {\rm EBIC}_\lambda=n\log({\rm RSS}_\lambda/n)+k_\lambda\log n
    +2\gamma\log \binom{p}{k_\lambda},
    \label{eq:ebic}
\end{equation}
where $n$ is the number of derivative-regression samples (the length of the target vector $\boldsymbol{d}$) and $\gamma=0.5$ by default. For interpolated data, these samples are correlated, so $n$ is the regression-sample count used by the criterion rather than an independent-sample count. In the model-selection experiment, the PDE simulations use $n_x=48$ and base $\Delta t=0.01$; derivative-regression rows are capped at 5000, and $n$ is the retained row count after this cap. The EBIC curve is used only for within-equation support-size selection, not for cross-method absolute likelihood comparison. The additional combinatorial penalty follows the EBIC rationale for large model spaces \citep{Chen2008}. Duplicate supports are collapsed by retaining the candidate with lowest EBIC. The resulting support-size Pareto curve, $k \mapsto \min_{\lambda:\,k_\lambda=k} {\rm EBIC}_\lambda$, is used for elbow detection: the selected model is the support size after which additional terms give limited EBIC improvement. The implementation uses \texttt{kneed}'s Kneedle-style knee detector when available and falls back to a distance-to-chord elbow rule otherwise \citep{Satopaa2011}. This selection rule does not use the true support to select the model; true support is used only afterward for evaluation.

\subsection{Benchmark differential equation systems and evaluation metrics}

\begin{figure}[!t]
\centerline{\includegraphics[width=0.95\textwidth]{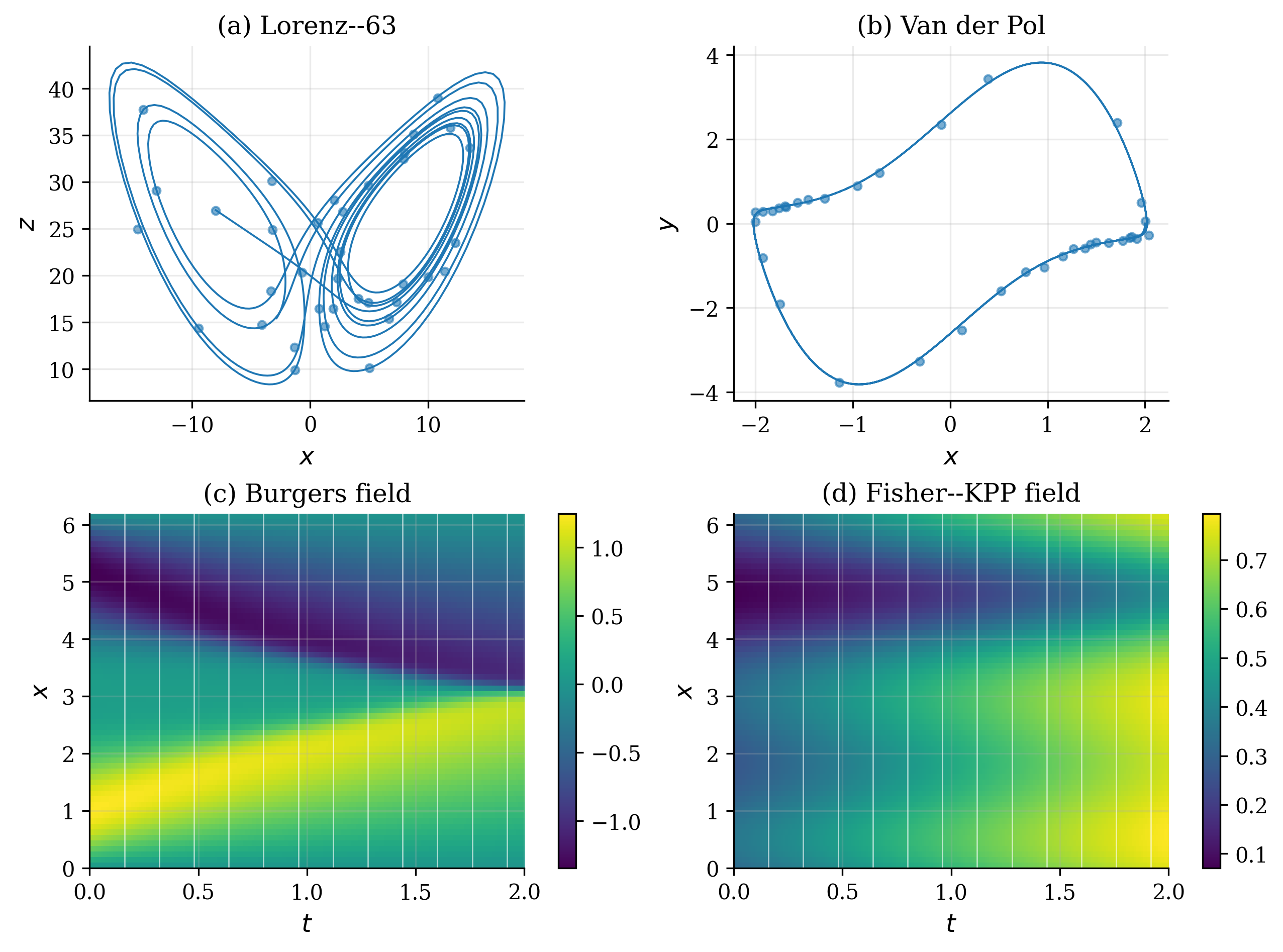}}
\caption{Representative trajectories and fields for Lorenz--63, Van der Pol, periodic Burgers, and Fisher--KPP. Markers in the ODE panels indicate sparse noisy observations at 3\% relative noise. White vertical markers in the PDE panels indicate sparse observation times; the heatmaps show the clean reference fields before noise injection}
\label{fig:datasets}
\end{figure}

Table~\ref{tab:systems} lists all systems used in the manuscript: two ODE benchmarks and three PDE benchmarks. Figure~\ref{fig:datasets} visualizes the two ODE systems and the two nonlinear PDE systems, and Figure~\ref{fig:advection_diffusion_dataset} visualizes the linear advection--diffusion PDE. Clean ODE trajectories are generated on a base grid with $\Delta t=0.005$, using $T=10$ for Lorenz--63 and $T=20$ for Van der Pol. The ODE setting covers sparse factors $\{8,16,32,64\}$, noise levels $\{0,0.01,0.03,0.05,0.10\}$, ten random seeds, and upsampling factor five for assisted methods. The main Burgers and Fisher--KPP PDE experiments use $n_x=64$, $\Delta t=0.005$, and $T=2$, with sparse factors $\{4,8,16,32\}$, the same noise levels, eight random seeds, and upsampling factor five. The advection--diffusion PDE uses $n_x=48$, $\Delta t=0.01$, $T=2$, sparse factors $\{4,8,16\}$, noise levels $\{0.01,0.03,0.05,0.10\}$, five random seeds, and upsampling factor five. The reported PDE runs use fixed system-specific POD ranks to limit high-rank noise amplification in Fisher--KPP while retaining suitable low-rank reconstructions for Burgers and advection--diffusion. All PDEs use periodic boundary conditions on $[0,2\pi)$. The main Fisher--KPP experiment uses $u_0(x)=0.25+0.10\sin x+0.08\cos(2x-0.3)$. The Fisher--KPP sensitivity and non-oracle model-selection experiment instead use the steeper periodic front-type initial condition $u_0(x)=0.25+0.18\tanh(2\sin x)+0.04\cos(2x)$, clipped below at 0.02, with the same diffusion coefficient $D=0.03$ and reaction coefficient $\rho=1$. Gaussian noise is scaled from the clean subsampled observations: ODE noise is componentwise, and PDE noise uses a single global field standard deviation. Temporal sparsity is imposed by subsampling the clean numerical solution and adding noise only at observed times. Temporal derivatives are computed by central finite differences on interior time points after any preprocessing, so the first and last time points are discarded in the derivative regression.

Support recovery is measured by the F1 score between the discovered and true supports. Let TP, FP, and FN denote the numbers of true-positive, false-positive, and false-negative active library terms. Then
\begin{equation}
    F_1=\frac{2\,\mathrm{TP}}{2\,\mathrm{TP}+\mathrm{FP}+\mathrm{FN}}.
    \label{eq:f1_score}
\end{equation}
Coefficient accuracy is measured by relative coefficient error, denoted ``Coeff. err.'' in the tables. A practical combined score is defined as
\begin{equation}
    S = \frac{F_1}{1+E_{\rm coeff}},
    \label{eq:practical_score}
\end{equation}
so that the diagnostic combines support recovery and coefficient accuracy. This score is paper-specific and applies to synthetic experiments with known coefficients. In the result tables, the column heading ``Score'' denotes this practical score $S$.

\begin{table}[!t]
\centering
\TBL{\caption{Synthetic systems used in the manuscript. Lorenz--63 and Van der Pol are the ODE benchmarks; periodic Burgers, Fisher--KPP, and periodic advection--diffusion are the PDE benchmarks.\label{tab:systems}}}
{\begin{tabular}{@{}lll@{}}\toprule
\TCH{Setting} & \TCH{System} & \TCH{Governing equation(s)} \\\midrule
\multirow{2}{*}{\makecell{ODE}} & Lorenz-63 & $\dot{x}=10(y-x)$, $\dot{y}=x(28-z)-y$, $\dot{z}=xy-8z/3$ \\
& Van der Pol & $\dot{x}=y$, $\dot{y}=2(1-x^2)y-x$ \\
\midrule
\multirow{3}{*}{\makecell{PDE}} & Burgers & $u_t=-u u_x+0.05u_{xx}$ \\
& Fisher--KPP & $u_t=0.03u_{xx}+u(1-u)$ \\
& Advection--diffusion & $u_t=-u_x+0.02u_{xx}$ \\\botrule
\end{tabular}}
\end{table}

\section{Empirical results}

The accuracy metrics in Tables~\ref{tab:ode_system} and \ref{tab:pde_system} are oracle-tuned over the STLSQ threshold grid described in Section~\ref{sec:methods}: the true support and coefficients are used to choose the best threshold on each synthetic record. The reported support F1 score, coefficient error, and practical score are therefore diagnostic best-case-over-grid quantities. They should be read as controlled measures of preprocessing quality.

\subsection{Ordinary differential equations}

Table~\ref{tab:ode_system} reports the ODE benchmark results for Lorenz--63 and Van der Pol. Polynomial EDMD reduces the median coefficient error for Lorenz--63 from 0.911 to 0.526 and for Van der Pol from 0.107 to 0.015, while also increasing the mean practical score on both systems. RBF EDMD is less accurate than polynomial EDMD for Lorenz--63 by coefficient error, but it still improves the practical score; on Van der Pol, it substantially reduces coefficient error. The support-F1 changes are smaller because the baseline often identifies the active terms but estimates their coefficients less accurately.

\begin{table}[H]
\centering
\TBL{\caption{ODE results for the Lorenz--63 and Van der Pol benchmark systems. The table is computed from the ODE parameter grid with sparse factors $\{8,16,32,64\}$, noise levels $\{0,0.01,0.03,0.05,0.10\}$, ten seeds, and upsampling factor five for EDMD methods. F1 and Score are means; Coeff. err. is the median over valid evaluation records.\label{tab:ode_system}}}
{\begin{tabular}{@{}llccc@{}}\toprule
\TCH{System} & \TCH{Method} & \TCH{F1} & \TCH{Coeff. err.} & \TCH{Score} \\\midrule
\multirow{3}{*}{\makecell[l]{Lorenz-63}} & Baseline & 0.725 & 0.911 & 0.477 \\
& EDMD-polynomial & \textbf{0.737} & \textbf{0.526} & \textbf{0.561} \\
& EDMD-RBF & 0.705 & 0.932 & 0.518 \\
\midrule
\multirow{3}{*}{\makecell[l]{Van der Pol}} & Baseline & 0.963 & 0.107 & 0.840 \\
& EDMD-polynomial & \textbf{0.999} & \textbf{0.015} & \textbf{0.977} \\
& EDMD-RBF & 0.983 & 0.019 & 0.936 \\\botrule
\end{tabular}}
\end{table}

Across matched ODE records, polynomial EDMD improves the practical score in 80.3\% of cases and the coefficient error in 80.8\%. RBF EDMD improves each quantity in 74.3\% of matched cases. These rates reinforce the conclusion: EDMD-based interpolation is most valuable for coefficient recovery, especially when the observable dictionary reflects the polynomial structure of the target equations.

\FloatBarrier
\subsection{Partial differential equations}

Table~\ref{tab:pde_system} reports the PDE benchmark results for periodic Burgers, Fisher--KPP, and advection--diffusion. The three PDEs respond differently to low-rank reconstruction. For Burgers, both assisted reconstructions improve coefficient accuracy and practical score: the baseline median coefficient error is 0.298, compared with 0.179 for optDMD and 0.217 for POD-EDMD-RBF. For Fisher--KPP with the default periodic initial condition, all three methods have coefficient errors near 0.039. The raw baseline is therefore highly competitive on this dataset, while optDMD gives the highest mean F1 and practical score on its valid records. For linear advection--diffusion, both optDMD and POD-EDMD-RBF substantially reduce the median coefficient error, from 0.166 for the baseline to 0.029 and 0.039, respectively. POD-EDMD-RBF gives the highest mean F1 and practical score on this system. Unlike the ODE benchmark, the PDE evidence supports a dynamics-dependent preprocessing benefit rather than a uniform improvement across equation systems.

The Fisher--KPP rows in Table~\ref{tab:pde_system} use the main initial condition stated above. To determine whether this limited assisted gain reflects the governing equation or the information content of the generated transient, we performed a sensitivity check using the steeper periodic front-type initial condition $u_0(x)=0.25+0.18\tanh(2\sin x)+0.04\cos(2x)$ and POD rank 2 while keeping the equation coefficients unchanged. With 1\% noise, sparse factor 8, upsampling factor 5, and five seeds, the baseline achieved mean F1 $0.800$, median coefficient error $0.043$, and mean score $0.767$. POD-EDMD-RBF achieved mean F1 $0.960$, median coefficient error $0.017$, and mean score $0.941$. The sensitivity result shows that Fisher--KPP can benefit from low-rank DMD-assisted preprocessing when the transient contains sharper spatial structure.

\begin{table}[H]
\centering
\TBL{\caption{PDE results for periodic Burgers, Fisher--KPP, and periodic linear advection--diffusion. The Burgers and Fisher--KPP rows are computed from the main PDE benchmark grid with $n_x=64$, $\Delta t=0.005$, $T=2$, sparse factors $\{4,8,16,32\}$, noise levels $\{0,0.01,0.03,0.05,0.10\}$, eight seeds, and upsampling factor five for DMD-assisted methods. The advection--diffusion rows are computed with $n_x=48$, $\Delta t=0.01$, $T=2$, sparse factors $\{4,8,16\}$, noise levels $\{0.01,0.03,0.05,0.10\}$, five seeds, upsampling factor five for optDMD and POD-EDMD-RBF, and POD rank four. F1 and Score are means; Coeff. err. is the median over valid evaluation records under the prescribed computational budget.\label{tab:pde_system}}}
{\begin{tabular}{@{}llccc@{}}\toprule
\TCH{System} & \TCH{Method} & \TCH{F1} & \TCH{Coeff. err.} & \TCH{Score} \\ \midrule
\multirow{3}{*}{\makecell[l]{Burgers}} & Baseline & 0.730 & 0.298 & 0.580 \\
& optDMD & \textbf{0.786} & \textbf{0.179} & \textbf{0.688} \\
& POD-EDMD-RBF & 0.775 & 0.217 & 0.658 \\
\midrule
\multirow{3}{*}{\makecell[l]{Fisher--KPP}} & Baseline & 0.837 & \textbf{0.039} & 0.810 \\
& optDMD & \textbf{0.846} & \textbf{0.039} & \textbf{0.822} \\
& POD-EDMD-RBF & 0.826 & \textbf{0.039} & 0.785 \\
\midrule
\multirow{3}{*}{\makecell[l]{Advection--\\diffusion}} & Baseline & 0.656 & 0.166 & 0.549 \\
& optDMD & 0.667 & \textbf{0.029} & 0.633 \\
& POD-EDMD-RBF & \textbf{0.703} & 0.039 & \textbf{0.656} \\ \botrule
\end{tabular}}
\end{table}

\begin{figure}[!t]
\centerline{\includegraphics[width=0.95\textwidth]{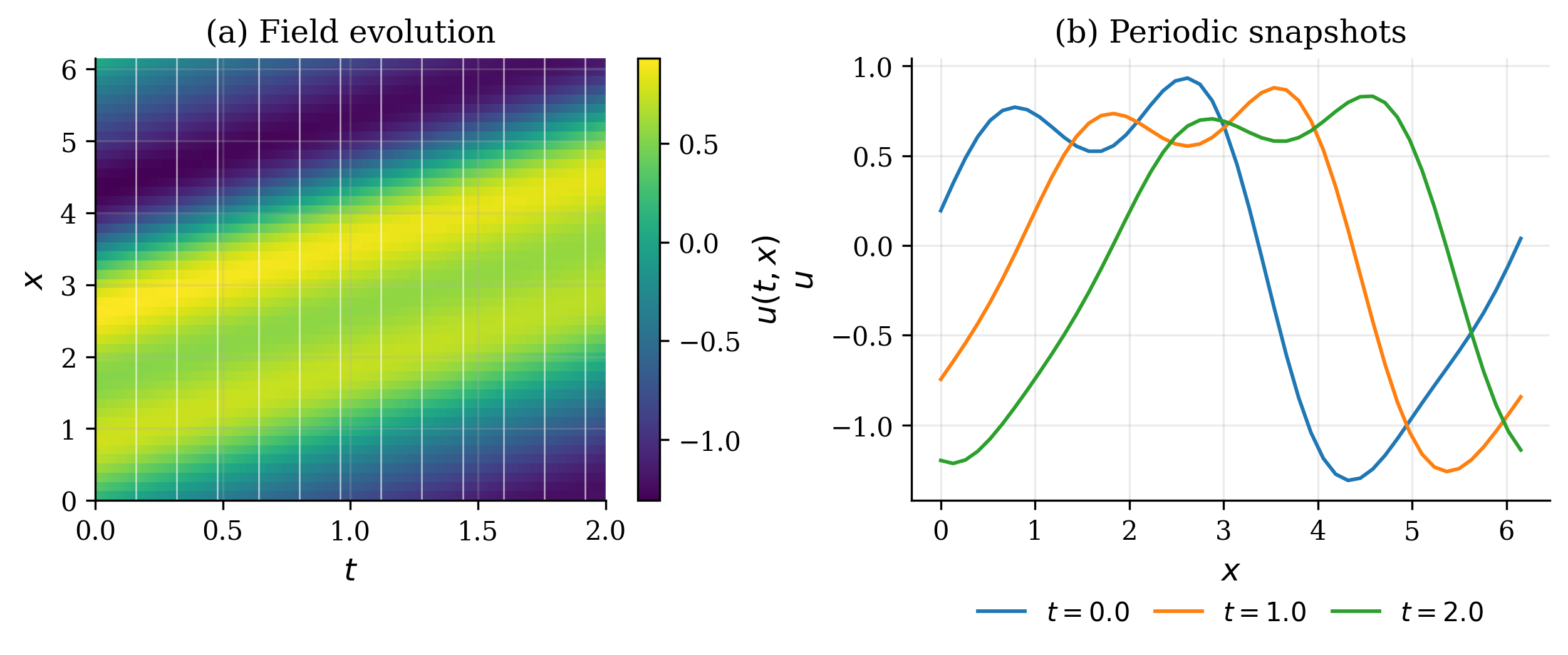}}
\caption{Representative field for the periodic advection--diffusion benchmark, $u_t=-u_x+0.02u_{xx}$. White vertical markers in the left panel indicate sparse observation times; the heatmap shows the clean reference field before noise injection. The right panel shows three periodic snapshots}
\label{fig:advection_diffusion_dataset}
\end{figure}

The periodic linear advection--diffusion benchmark is
\[
    u_t=-c u_x+\nu u_{xx},\qquad c=1,
    \quad \nu=0.02,
\]
with a multi-frequency periodic initial condition. This system complements the nonlinear Burgers and Fisher--KPP tests by adding a linear PDE whose temporal evolution is generated by a linear spatial differential operator, while PDE-FIND must still identify the active library terms $u_x$ and $u_{xx}$ from noisy sparse observations. Figure~\ref{fig:advection_diffusion_dataset} visualizes the clean field and representative snapshots. Across matched records, POD-EDMD-RBF improves the advection--diffusion coefficient error in 98.3\% of cases and the practical score in 96.7\% of cases; optDMD improves each quantity in 83.3\% of cases. At noise levels $0.01,0.03,0.05,$ and $0.10$, the baseline median errors are $0.025,0.103,0.227,$ and $0.545$, POD-EDMD-RBF gives $0.017,0.031,0.057,$ and $0.199$, and optDMD gives $0.028,0.027,0.029,$ and $0.120$. This pattern supports the interpretation that DMD-assisted upsampling is most effective when the low-rank temporal reconstruction is compatible with the PDE evolution.

The PDE benchmark therefore separates three cases. In periodic Burgers, both assisted low-rank reconstructions improve practical score and coefficient error over the baseline. In the main Fisher--KPP dataset, the raw baseline is competitive; optDMD gives a modest practical-score gain on valid records, whereas POD-EDMD-RBF is less favorable. In periodic advection--diffusion, both assisted low-rank reconstructions improve coefficient error, with optDMD giving the lowest median coefficient error and POD-EDMD-RBF giving the highest practical score. Across matched records in the Burgers and Fisher--KPP benchmark grid, POD-EDMD-RBF improves practical score in 64.7\% of cases and coefficient error in 64.4\% of cases. optDMD improves each quantity in 70.4\% of matched valid records under the prescribed computational budget.

\begin{figure}[!t]
\centerline{\includegraphics[width=0.95\textwidth]{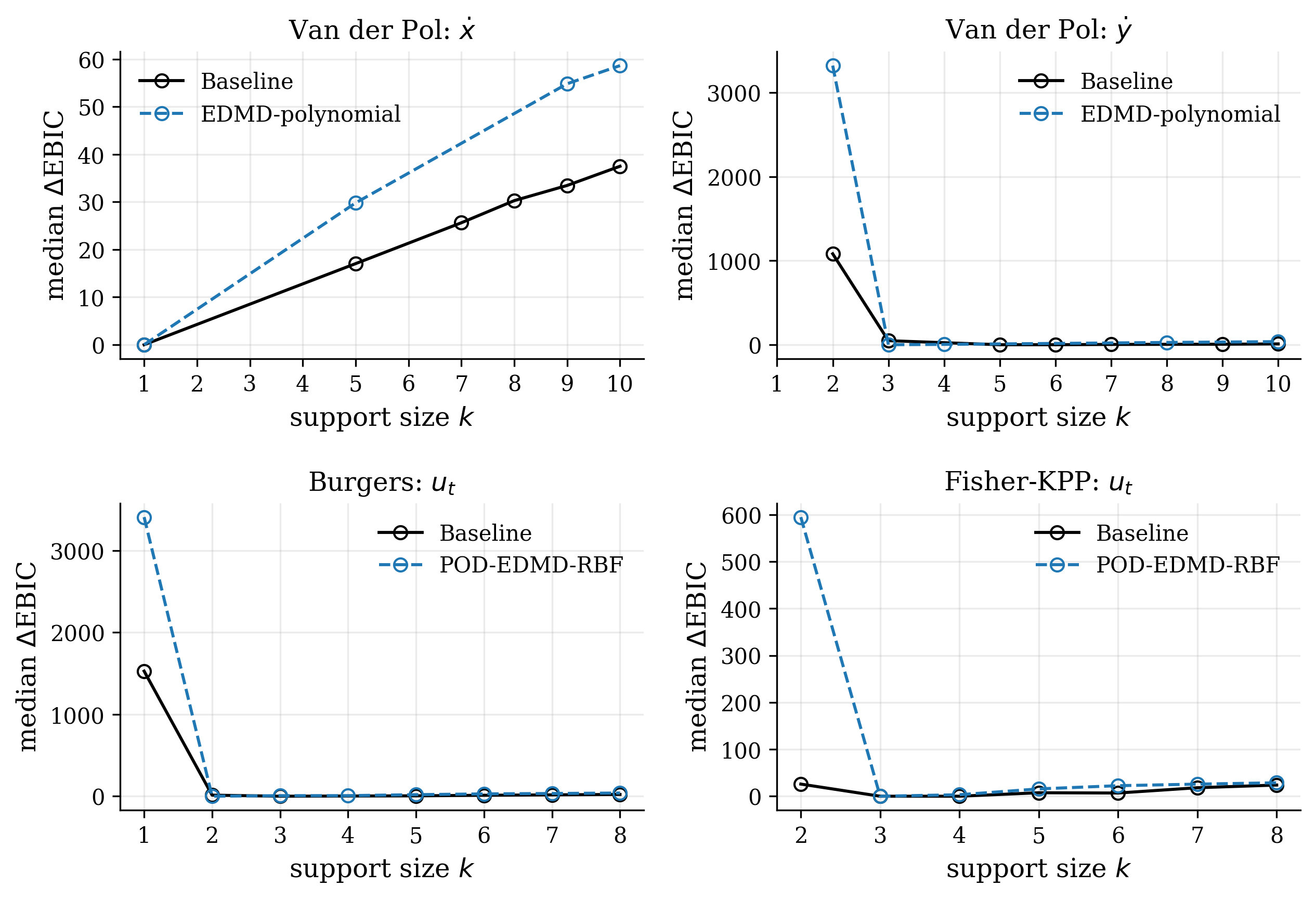}}
\caption{Pareto plots for non-oracle model selection on Van der Pol, Burgers, and Fisher--KPP with a steep periodic front-type initial condition. Each curve summarizes the median $\Delta$EBIC over seeds after collapsing duplicate supports by support size. The selected model is chosen by elbow detection on the EBIC/support-size curve, not by the true support}
\label{fig:model_selection_pareto}
\end{figure}

Figure~\ref{fig:model_selection_pareto} shows the corresponding equation-wise Pareto plots. Table~\ref{tab:model_selection} focuses on the default DMD-assisted selector because this experiment asks whether upsampling makes support-size selection reliable without access to true coefficients. Under the same elbow rule, the raw baseline is mostly but not uniformly successful: it selects the correct support size for Van der Pol $\dot{x}$, Van der Pol $\dot{y}$, and Burgers $u_t$ in all five seeds. For Fisher--KPP with a steep periodic front-type initial condition, the baseline selects the correct median support size $k=3$, but only three of five seeds select the exact support size; those three seeds choose the wrong support $\{1,u,u^2\}$, replacing $u_{xx}$ with an intercept, while one seed under-selects and one over-selects. Thus the baseline never recovers the true Fisher--KPP support in any seed, giving a mean support F1 score of 0.660. In contrast, the DMD-assisted selector recovers the exact support $\{u,u^2,u_{xx}\}$ in all five Fisher--KPP seeds and selects the true support size for every target equation in every seed. Median coefficient errors also decrease under the assisted reconstructions, from 0.005 to 0.001 for Van der Pol $\dot{x}$, 0.031 to 0.005 for Van der Pol $\dot{y}$, 0.046 to 0.020 for Burgers, and 0.059 to 0.011 for Fisher--KPP.

\FloatBarrier
\begin{table}[H]
\centering
\TBL{\caption{Non-oracle EBIC/Pareto model selection for the Van der Pol ODE equations and the Burgers and Fisher--KPP PDE equations, with Fisher--KPP evaluated under a steep periodic front-type initial condition, after DMD-assisted upsampling. ``Non-oracle'' means that the elbow detector selects the support size from the EBIC/support-size curve without using the true support or true coefficients; ground truth is used only afterward to verify the selected support size. The experiment uses 1\% noise, sparse factor 8 for Van der Pol, sparse factor 4 for PDEs, upsampling factor 5, POD rank 2 for Fisher--KPP, five random seeds, $n_x=48$ and base $\Delta t=0.01$ for PDEs, and at most 5000 derivative-regression rows for EBIC evaluation. Selected $k$ is reported separately for each target equation; all rows select the shown value in all five seeds.\label{tab:model_selection}}}
{\begin{tabular}{@{}lllccc@{}}\toprule
\TCH{Setting} & \TCH{System} & \TCH{Target variable} & \TCH{Method} & \TCH{True $k$} & \TCH{Selected $k$} \\ \midrule
ODE & Van der Pol & $\dot{x}$ & EDMD-polynomial & 1 & 1 \\
 & Van der Pol & $\dot{y}$ & EDMD-polynomial & 3 & 3 \\
\midrule
PDE & Burgers & $u_t$ & POD-EDMD-RBF & 2 & 2 \\
 & Fisher--KPP & $u_t$ & POD-EDMD-RBF & 3 & 3 \\ \botrule
\end{tabular}}
\end{table}

\subsection{Interpolation-strategy ablation}
\label{subsec:strategy_ablation}

The main benchmark evaluates EDMD and POD-EDMD-RBF with local-reset interpolation, because it is simple, uses each observed snapshot as an anchor, and avoids the long-horizon drift that can occur when a learned linear observable model is rolled forward from the initial time ($t_1$) over the entire window. Appendix~\ref{app:upsampling_strategies} reports an ablation study of two nearby alternatives: residual-corrected rollout and two-sided interpolation. The ablation supports keeping local reset as the main default while observing that residual correction can reduce ODE coefficient error and two-sided interpolation can give marginal PDE gains in the most sparse/noisy setting. Appendix~\ref{app:classical_interpolation} adds a separate comparison against non-dynamical interpolation techniques: linear interpolation and smoothing splines are tuned only from the sparse noisy observations and are then evaluated with the same downstream discovery procedure.

\section{Discussion}

The results support the central hypothesis that dynamics-aware upsampling can improve sparse equation discovery from noisy, sparse measurements. For the tested ODE problems, the main benefit appears in coefficient accuracy. Polynomial EDMD is the strongest ODE preprocessor because the benchmarked ODEs have polynomial right-hand sides and the EDMD dictionary is aligned with the SINDy library. Appendix~\ref{app:classical_interpolation} shows that linear interpolation and observation-tuned smoothing also improve over the raw sparse baseline, while EDMD-polynomial retains the best ODE coefficient error and practical score.

For PDE-FIND, the gains are system-dependent. Low-rank DMD-assisted preprocessing acts as joint temporal interpolation and spatial POD denoising: the baseline differentiates the raw noisy field, whereas the POD methods differentiate a rank-$r$ reconstruction. POD-EDMD-RBF is the most reliable selected PDE preprocessor across the main benchmark. Optimized DMD is a competitive low-rank alternative. Appendix~\ref{app:classical_interpolation} shows that POD-EDMD-RBF gives the best aggregate PDE support F1, coefficient error, and practical score against linear and smoothing-spline interpolation techniques, with a modest margin over the tuned spline. These results indicate that DMD-assisted preprocessing adds value when the field evolution is compatible with the learned low-rank representation.

The oracle-threshold procedure is a controlled diagnostic, not a deployable discovery rule. It measures whether a preprocessing method gives STLSQ a cleaner regression problem when the best support-aware threshold is chosen from a fixed grid. It does not imply that this threshold would be known for experimental data. The EBIC/Pareto experiment partially addresses this limitation by selecting along the threshold path without using true support labels. This design is closer to practical sparse-equation selection because it asks whether DMD-assisted preprocessing makes the complexity--fit elbow easier to identify. In experimental use, upsampling should still be coupled with validation residuals, stability selection, information criteria, ensemble stability checks, uncertainty-aware selection, or multi-criteria support decision \citep{Mangan2017,Fasel2022,Kaptanoglu2022,Thanasutives2024UBIC,Thanasutives2026KOPDE}.

The selected methods are chosen to probe two specific mechanisms: nonlinear observable lifting for ODEs and low-rank temporal reconstruction for PDEs. Other DMD variants may be useful for specialized data regimes, but a broader algorithmic survey is outside the scope of this paper.

\section{Limitations and future work}

The benchmark is synthetic, so true supports and true coefficients are available for oracle scoring. In experimental settings, oracle threshold selection should be replaced by non-oracle procedures such as the EBIC/Pareto workflow demonstrated here, validation residuals, information criteria, stability selection, or a carefully designed best-subset or mixed-integer formulation when the library is small enough. The PDE examples are periodic and smooth, allowing spectral spatial derivatives. More difficult PDEs with shocks, nonperiodic boundaries, high-order derivatives, irregular sensors, strong candidate-term collinearity, or variable coefficients likely require weak-form SINDy, integral formulations, uncertainty-aware criteria, knockoff-based false-discovery controls, or additional spatial regularization \citep{Schaeffer2013,Schaeffer2017,Messenger2021,Thanasutives2023nPIML,Thanasutives2024UBIC,Thanasutives2025VBIC,Thanasutives2026KOPDE}. Appendix~\ref{app:qr_sensitivity} includes the $q=1$ POD-only denoising control and the $q>1$ POD-plus-upsampling cases, showing that the spatial and temporal components of the PDE preprocessing can be inspected separately in a representative sensitivity setting. Appendix~\ref{app:classical_interpolation} focuses on direct linear and smoothing-spline techniques. Broader comparisons with state-space smoothers, Gaussian processes, additional interpolation variants, and other DMD variants are left for future work. Accuracy metrics are computed over valid evaluation records under the prescribed computational budget.

\section{Conclusion}

This paper presents a comprehensive benchmark for Koopman-based upsampling before sparse equation discovery. The evaluated method set contains the no-upsampling baseline and two assisted preprocessors per setting: polynomial EDMD and RBF EDMD for ODEs, and optimized DMD and POD-EDMD-RBF for PDEs. The ODE evidence is strongest for polynomial EDMD, which substantially reduces coefficient error and improves the practical score. The PDE evidence is system-dependent, but POD-EDMD-RBF is the most reliable selected PDE preprocessor across the full grid, and optimized DMD is a competitive low-rank alternative on the valid subset allowed by the computational budget. A comparison with observation-tuned linear and smoothing-spline interpolation techniques shows that the selected Koopman-based preprocessors retain the best aggregate ODE and PDE performance. Overall, Koopman-based upsampling is most useful when the observable dictionary, low-rank representation, and interpolation strategy are matched to the dynamics and sampling regime.

\appendix

\section{Upsampling-factor and POD-rank sensitivity}
\label{app:qr_sensitivity}

This appendix examines the fixed interpolation factor $q$ and POD rank $r$ used by the assisted preprocessors. For PDEs, the $q=1$ column is a POD-only denoising control: the noisy sparse snapshots are projected to rank $r$ and reconstructed at the observed times, but no temporal points are inserted. Values with $q>1$ combine the same POD projection with DMD/EDMD temporal upsampling. The study is not used for oracle selection in the main tables; it is a representative robustness check at the same synthetic-equation level. Unless otherwise stated, the sensitivity run uses relative noise 0.03, ODE sparse factor 16, PDE sparse factor 8, and 5 random seeds.

\begin{figure}[H]
\centering
\includegraphics[width=0.95\textwidth]{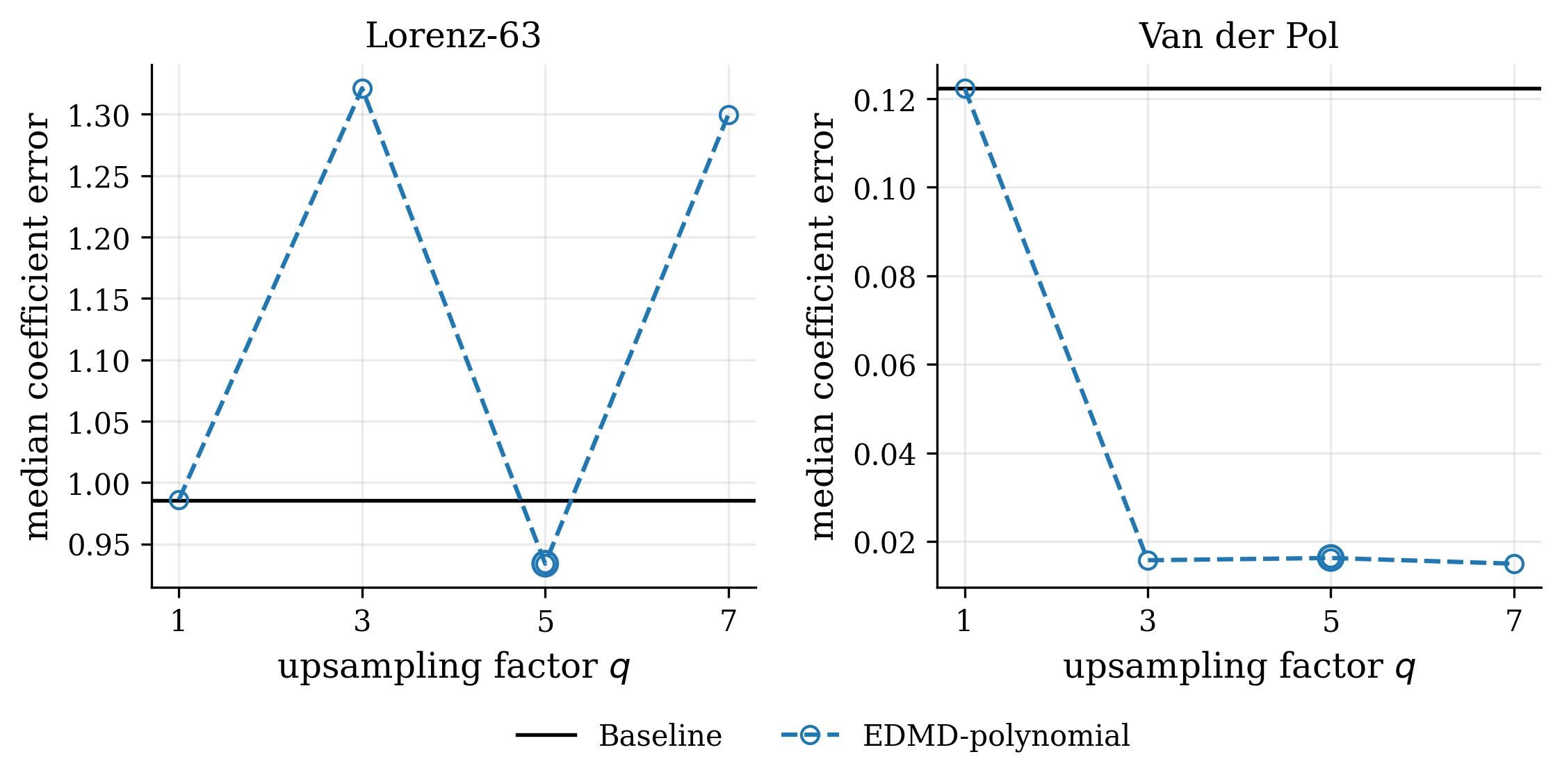}
\caption{ODE sensitivity to the upsampling factor $q$ for EDMD-polynomial preprocessing. Horizontal black lines show the no-upsampling baseline; dashed blue curves show EDMD-polynomial. Markers at $q=5$ indicate the value used in the main benchmark.}
\label{fig:appendix_q_sensitivity_ode}
\end{figure}

Figure~\ref{fig:appendix_q_sensitivity_ode} and Table~\ref{tab:appendix_ode_q_sensitivity} show that the usefulness of increasing $q$ is system-dependent. For Van der Pol, EDMD-polynomial is already substantially better than the baseline for $q\geq3$, and the default $q=5$ is near the best observed value. For Lorenz--63, the sensitivity is less monotone, which is consistent with the chaotic trajectory and the difficulty of estimating accurate derivatives from sparse noisy measurements. The default $q=5$ gives the lowest median coefficient error among the EDMD-polynomial values tested for Lorenz--63, although the practical score varies only modestly across the grid.

\begin{table}[H]
\centering
\TBL{\caption{ODE $q$ sensitivity for Lorenz--63 and Van der Pol under EDMD-polynomial preprocessing. The sensitivity run uses relative noise 0.03, sparse factor 16, and five random seeds. F1 and Score are means; Coeff. err. is the median over valid evaluation records. Boldface marks the best value at reported precision within each system.\label{tab:appendix_ode_q_sensitivity}}}
{\begin{tabular}{@{}llcccc@{}}\toprule
\TCH{System} & \TCH{Method} & \TCH{$q$} & \TCH{F1} & \TCH{Coeff. err.} & \TCH{Score} \\ \midrule
Lorenz--63 & Baseline & -- & 0.554 & 0.986 & 0.268 \\
 & EDMD-polynomial & 1 & 0.562 & 0.986 & 0.272 \\
 & EDMD-polynomial & 3 & 0.591 & 1.321 & 0.261 \\
 & EDMD-polynomial & 5 & 0.563 & \textbf{0.934} & \textbf{0.288} \\
 & EDMD-polynomial & 7 & \textbf{0.609} & 1.300 & 0.273 \\
\midrule
Van der Pol & Baseline & -- & \textbf{1.000} & 0.122 & 0.893 \\
 & EDMD-polynomial & 1 & \textbf{1.000} & 0.122 & 0.893 \\
 & EDMD-polynomial & 3 & \textbf{1.000} & 0.016 & 0.982 \\
 & EDMD-polynomial & 5 & \textbf{1.000} & 0.016 & \textbf{0.984} \\
 & EDMD-polynomial & 7 & \textbf{1.000} & \textbf{0.015} & \textbf{0.984} \\
\botrule
\end{tabular}}
\end{table}

\begin{figure}[H]
\centering
\includegraphics[width=0.95\textwidth]{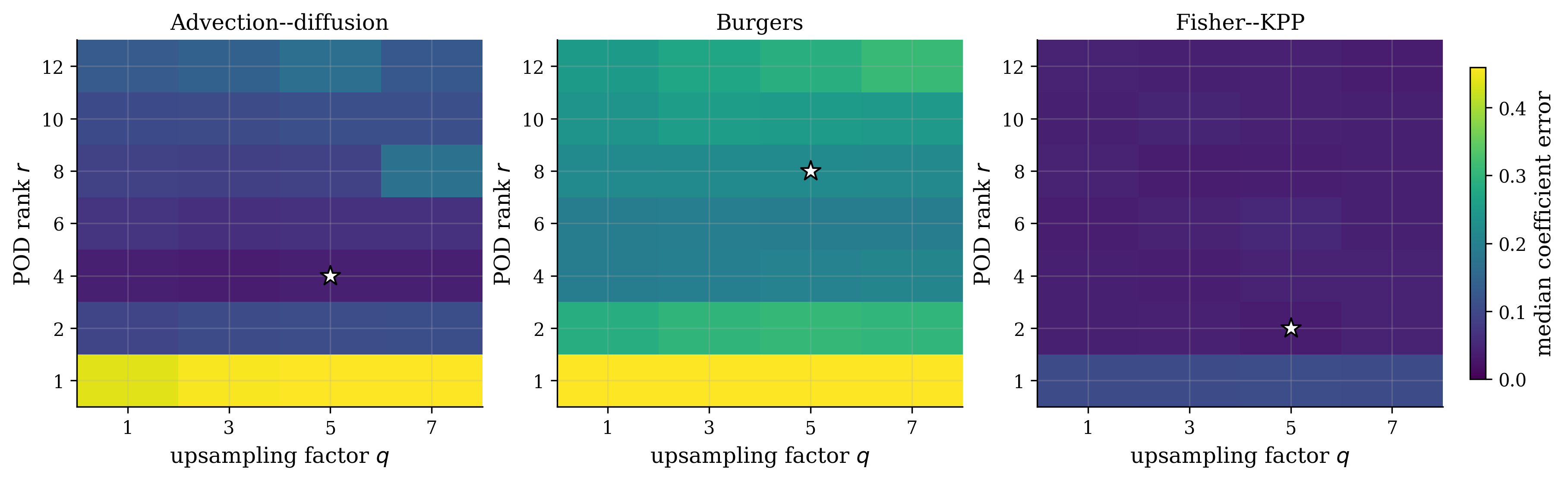}
\caption{PDE sensitivity of POD-EDMD-RBF preprocessing to the interpolation factor $q$ and POD rank $r$. Each heatmap reports median coefficient error at relative noise 0.03 and PDE sparse factor 8; white stars mark the default values used in the main benchmark. The $q=1$ column corresponds to POD-only denoising without temporal insertion.}
\label{fig:appendix_qr_sensitivity_pde}
\end{figure}

\begin{table}[H]
\centering
\TBL{\caption{Default-versus-best comparison over the PDE $q$--$r$ sensitivity grid for advection--diffusion, Burgers, and Fisher--KPP. The best grid point is selected by mean Score, with Coeff. err. used only as a tie-breaker. F1 and Score are means; Coeff. err. is the median over five seeds. Boldface marks the better value between the default and best grid point at reported precision.\label{tab:appendix_pde_qr_sensitivity}}}
{%
\begin{tabular}{@{}lcccccccc@{}}\toprule
\TCH{System} & \multicolumn{4}{c}{\TCH{Default}} & \multicolumn{4}{c}{\TCH{Best}} \\
\cmidrule(lr){2-5}\cmidrule(lr){6-9}
 & \TCH{$(q,r)$} & \TCH{F1} & \TCH{Coeff. err.} & \TCH{Score} & \TCH{$(q,r)$} & \TCH{F1} & \TCH{Coeff. err.} & \TCH{Score} \\ \midrule
Advection--diffusion & (5,4) & \textbf{0.667} & 0.041 & 0.641 & (3,4) & \textbf{0.667} & \textbf{0.035} & \textbf{0.643} \\
Burgers & (5,8) & \textbf{0.667} & 0.220 & 0.545 & (1,6) & \textbf{0.667} & \textbf{0.193} & \textbf{0.559} \\
Fisher--KPP & (5,2) & \textbf{0.800} & \textbf{0.037} & \textbf{0.771} & (7,10) & \textbf{0.800} & 0.041 & \textbf{0.771} \\
\botrule
\end{tabular}%
}
\end{table}

The PDE grid in Figure~\ref{fig:appendix_qr_sensitivity_pde} and Table~\ref{tab:appendix_pde_qr_sensitivity} supports the use of fixed system-specific ranks in the main study. The $q=1$ column provides the requested POD-only control, while $q>1$ combines the same low-rank spatial projection with temporal upsampling. The default advection--diffusion choice $(q,r)=(5,4)$ is close to the best grid point $(3,4)$; the Burgers default $(5,8)$ is not the lowest-error point in this sensitivity slice but remains in a stable region of the heatmap; and Fisher--KPP has a broad low-error region in which the default rank $r=2$ remains competitive. These results do not imply that $q$ and $r$ are universally optimal. They show that the main conclusions are not based on an isolated unstable choice of interpolation factor or POD rank.

\section{Alternative DMD interpolation strategies}
\label{app:upsampling_strategies}

This appendix reports a targeted ablation of how the same fitted EDMD or POD-EDMD-RBF model is evaluated between observed snapshots. The stress setting uses the highest noise level and sparsest sampling in the benchmark: noise level $0.10$, ODE sparse factor $64$, PDE sparse factor $32$, upsampling factor $q=5$, and 20 random seeds. Global rollout was tested in the script but is omitted from Table~\ref{tab:upsampling_strategy_app} because it consistently underperforms under the ODE setting. The two alternatives reported here are residual-corrected rollout and two-sided interpolation. Residual correction fits a global rollout and adds a smooth correction through the observation residuals at the anchor times. Two-sided interpolation blends forward evolution from the left endpoint and backward evolution from the right endpoint in each observed interval.

\begin{table}[H]
\centering
\TBL{\caption{Stress-condition ablation for alternative DMD/EDMD interpolation strategies on the ODE and PDE benchmark groups. The ODE rows aggregate Lorenz--63 and Van der Pol; the PDE rows aggregate periodic Burgers and Fisher--KPP under the stress setting with 10\% noise, ODE sparse factor 64, PDE sparse factor 32, upsampling factor five, and 20 seeds. Local reset is the main-paper default. Lower Coeff. err. and higher Score are better. Boldface marks the best interpolation strategy within each setting; the no-upsampling baseline is shown for reference.\label{tab:upsampling_strategy_app}}}
{%
\begin{tabular}{llccc}
\toprule
\TCH{Setting/method} & \TCH{Upsampling strategy} & \TCH{F1} & \TCH{Coeff. err.} & \TCH{Score} \\
\midrule
\multirow{4}{*}{\makecell[l]{ODE\\EDMD-polynomial}}
& Baseline, no upsampling & 0.805 & 0.703 & 0.522 \\
& Local reset (main paper) & 0.760 & 0.681 & \textbf{0.569} \\
& Residual-corrected rollout & \textbf{0.767} & \textbf{0.575} & 0.566 \\
& Two-sided interpolation & 0.719 & 0.799 & 0.491 \\
\midrule
\multirow{4}{*}{\makecell[l]{ODE\\EDMD-RBF}}
& Baseline, no upsampling & 0.805 & 0.703 & 0.522 \\
& Local reset (main paper) & 0.715 & 0.849 & 0.509 \\
& Residual-corrected rollout & \textbf{0.749} & \textbf{0.571} & \textbf{0.555} \\
& Two-sided interpolation & 0.691 & 0.983 & 0.447 \\
\midrule
\multirow{4}{*}{\makecell[l]{PDE\\POD-EDMD-RBF}}
& Baseline, no upsampling & 0.733 & 0.409 & 0.571 \\
& Local reset (main paper) & 0.707 & 0.457 & 0.555 \\
& Residual-corrected rollout & 0.680 & 0.613 & 0.522 \\
& Two-sided interpolation & \textbf{0.711} & \textbf{0.450} & \textbf{0.558} \\
\botrule
\end{tabular}%
}
\end{table}

\clearpage
\section{Classical non-dynamical interpolation techniques}
\label{app:classical_interpolation}

This appendix evaluates whether the improvements from Koopman-based preprocessing can be explained by inserting additional time points before sparse regression. The comparison focuses on direct non-dynamical alternatives: the no-upsampling baseline, componentwise piecewise-linear interpolation, componentwise smoothing-spline interpolation, and the main assisted preprocessors. The ODE assisted method is EDMD-polynomial. The PDE assisted method is POD-EDMD-RBF.

The interpolation factor $q$ is fixed across all non-baseline methods rather than tuned by validation, because it controls the output grid resolution used for subsequent derivative estimation. Observation-only validation at the original sparse measurement times cannot identify a preferred dense-grid spacing without using downstream SINDy/PDE-FIND information. In contrast, the smoothing-spline parameter and the PDE POD rank $r$ directly affect reconstruction of held-out sparse noisy measurements, so these parameters are selected by deterministic interior holdout validation using only the observed sparse noisy trajectories or snapshots.

No dense clean trajectory, true coefficient vector, threshold oracle, or downstream SINDy/PDE-FIND score is used to tune any Appendix~C preprocessing parameter. The appendix therefore provides a fair check against the most direct alternative explanation: ordinary temporal interpolation plus classical smoothing. The Appendix-C advection--diffusion comparison follows this grid, which differs from the main advection--diffusion experiment reported in Table~\ref{tab:pde_system}.

\begin{table}[H]
\centering
\TBL{\caption{Classical non-dynamical interpolation techniques for the ODE benchmark. Linear interpolation and the validation-tuned smoothing spline insert the same number of temporal points as the assisted method but do not estimate a flow map or Koopman operator. The spline smoothing parameter is selected only from sparse noisy observations. F1 and Score are means, Coeff. err. is the median, and boldface marks the best value within each system and metric.\label{tab:appendix_c_ode_classical}}}
{%
\begin{tabular}{@{}llcccc@{}}
\toprule
\TCH{System} & \TCH{Method} & \TCH{$q$} & \TCH{F1} & \TCH{Coeff. err.} & \TCH{Score} \\
\midrule
Lorenz--63 & Baseline & -- & 0.725 & 0.911 & 0.477 \\
 & Linear interpolation & 5 & 0.727 & 0.911 & 0.493 \\
 & Tuned smoothing spline & 5 & 0.730 & 0.936 & 0.530 \\
 & EDMD-polynomial & 5 & \textbf{0.737} & \textbf{0.526} & \textbf{0.561} \\
\midrule
Van der Pol & Baseline & -- & 0.963 & 0.107 & 0.840 \\
 & Linear interpolation & 5 & 0.998 & 0.049 & 0.937 \\
 & Tuned smoothing spline & 5 & 0.995 & 0.027 & 0.957 \\
 & EDMD-polynomial & 5 & \textbf{0.999} & \textbf{0.015} & \textbf{0.977} \\
\botrule
\end{tabular}%
}
\end{table}

The ODE comparison in Table~\ref{tab:appendix_c_ode_classical} supports the main conclusion: non-dynamical interpolation improves over the raw sparse baseline, while EDMD-polynomial gives the best score and coefficient accuracy in both ODE systems.

\begin{table}[H]
\centering
\TBL{\caption{Classical non-dynamical interpolation techniques and observation-validated POD-DMD and Koopman-based preprocessing for the PDE benchmark. The linear and validation-tuned smoothing-spline techniques apply separate temporal interpolants at each spatial grid point, with a single validation-selected smoothing level shared across the field; they do not use POD, DMD, or EDMD. For POD-EDMD-RBF, the POD rank $r$ is selected by deterministic interior holdout validation on the sparse noisy snapshots only. All non-baseline methods use the same fixed interpolation factor $q$. F1 and Score are means, Coeff. err. is the median, and boldface marks the best value within each system and metric.\label{tab:appendix_c_pde_classical}}}
{%
\begin{tabular}{@{}llccccc@{}}
\toprule
\TCH{System} & \TCH{Method} & \TCH{$q$} & \TCH{$r$} & \TCH{F1} & \TCH{Coeff. err.} & \TCH{Score} \\
\midrule
Burgers & Baseline & -- & -- & 0.730 & 0.298 & 0.580 \\
 & Linear interpolation & 5 & -- & 0.732 & 0.252 & 0.598 \\
 & Tuned smoothing spline & 5 & -- & 0.752 & \textbf{0.165} & 0.661 \\
 & POD-EDMD-RBF & 5 & 4 & \textbf{0.788} & 0.192 & \textbf{0.692} \\
\midrule
Fisher--KPP & Baseline & -- & -- & 0.837 & 0.039 & 0.810 \\
 & Linear interpolation & 5 & -- & 0.839 & 0.039 & 0.812 \\
 & Tuned smoothing spline & 5 & -- & \textbf{0.857} & 0.042 & \textbf{0.831} \\
 & POD-EDMD-RBF & 5 & 2 & 0.851 & \textbf{0.036} & 0.825 \\
\midrule
Advection--diffusion & Baseline & -- & -- & 0.695 & 0.168 & 0.593 \\
 & Linear interpolation & 5 & -- & 0.702 & 0.118 & 0.615 \\
 & Tuned smoothing spline & 5 & -- & 0.722 & \textbf{0.024} & 0.687 \\
 & POD-EDMD-RBF & 5 & 4 & \textbf{0.734} & 0.027 & \textbf{0.695} \\
\botrule
\end{tabular}%
}
\end{table}

In the PDE comparison in Table~\ref{tab:appendix_c_pde_classical}, tuned smoothing splines are competitive, especially for coefficient accuracy in some rows, while POD-EDMD-RBF gives the best aggregate PDE practical score and support recovery over the same publication grid. Thus, the improvements in the main benchmark are not explained solely by adding intermediate time points. The PDE results indicate a system-dependent advantage for learned low-rank dynamics rather than a uniform dominance over classical smoothing.

\begin{Backmatter}

\paragraph{Abbreviations}
DMD, dynamic mode decomposition; EBIC, extended Bayesian information criterion; EDMD, extended dynamic mode decomposition; ODE, ordinary differential equation; PDE, partial differential equation; PDE-FIND, PDE functional identification; POD, proper orthogonal decomposition; RBF, radial basis function; RSS, residual sum of squares; SINDy, sparse identification of nonlinear dynamics; STLSQ, sequential thresholded least squares.


\paragraph{Funding Statement}
Pongpisit Thanasutives is supported by the research fund from the Special Postdoctoral Researcher
(SPDR) Program at RIKEN, Japan.

\paragraph{Competing Interests}
The author declares none.

\paragraph{Data Availability Statement}
The manuscript is accompanied by reproducible Python scripts and benchmark outputs, which are available online at \url{https://github.com/Pongpisit-Thanasutives/Koopman-SINDy}.

\paragraph{Ethical Standards}
This work does not involve human participants, animal subjects, or private data.

\paragraph{Author Contributions}
Conceptualization: Pongpisit Thanasutives (lead). Formal analysis: Pongpisit Thanasutives (lead). Funding acquisition: Yoshinobu Kawahara (lead), Pongpisit Thanasutives (supporting). Investigation: Pongpisit Thanasutives (lead). Methodology: Pongpisit Thanasutives (lead). Project administration: Yoshinobu Kawahara (lead). Resources: Yoshinobu Kawahara (lead). Software: Pongpisit Thanasutives (lead). Validation: Pongpisit Thanasutives (lead). Visualization: Pongpisit Thanasutives (lead). Writing--original draft: Pongpisit Thanasutives (lead), Yoshinobu Kawahara (supporting). Writing--review and editing: Pongpisit Thanasutives (lead), Yoshinobu Kawahara (supporting). Supervision: Yoshinobu Kawahara (lead).

\paragraph{Supplementary Material}
No supplementary material is available for this manuscript.

\bibliographystyle{plainnat}
\bibliography{references}

\end{Backmatter}

\end{document}